\documentclass[letterpaper]{article} 
\usepackage{aaai2026}  
\usepackage{times}  
\usepackage{helvet}  
\usepackage{courier}  
\usepackage[hyphens]{url}  
\usepackage{graphicx} 
\usepackage{natbib}  
\usepackage{caption} 
\usepackage{algorithm}
\usepackage{algorithmic}
\usepackage{newfloat}
\usepackage{listings}

\usepackage{amsmath}
\usepackage{amssymb}
\usepackage{graphicx} 
\usepackage{caption} 
\usepackage{dsfont}
\usepackage{tikz}
\usepackage{amsthm}
\usepackage{subfigure}
\usepackage{xcolor}
\usepackage{xr}

\newcommand{\unif}{\pazocal{U}}
\newcommand*\circled[1]{\tikz[baseline=(char.base)]{
            \node[shape=circle,draw,inner sep=2pt] (char) {#1};}}
\newtheorem{theorem}{Theorem}
\DeclareCaptionStyle{ruled}{labelfont=normalfont,labelsep=colon,strut=off} 
\floatstyle{ruled}
\newfloat{listing}{tb}{lst}{}
\floatname{listing}{Listing}
\title{Measuring Model Performance in the Presence of an Intervention}
\author{
    Winston Chen\textsuperscript{\rm 1}
    Michael W. Sjoding\textsuperscript{\rm 2},
    Jenna Wiens\textsuperscript{\rm 1}
}
\affiliations{
    \textsuperscript{\rm 1}Division of Computer Science and Engineering,  University of Michigan, Ann Arbor, MI 48109, USA\\
    \textsuperscript{\rm 2}Division of Pulmonary and Critical Care Medicine,  University of Michigan Medical School, Ann Arbor, MI 48109, USA\\
    chenwt@umich.edu, msjoding@med.umich.edu, wiensj@umich.edu 
}
\def\isMain{}
\begin{document}

\maketitle

\begin{abstract}
AI models are often evaluated based on their ability to predict the outcome of interest. 
However, in many AI for social impact applications, the presence of an intervention that affects the outcome can bias the evaluation. 
Randomized controlled trials (RCTs) randomly assign interventions, allowing data from the control group to be used for unbiased model evaluation. 
However, this approach is inefficient because it ignores data from the treatment group. 
Given the complexity and cost often associated with RCTs, making the most use of the data is essential. 
Thus, we investigate model evaluation strategies that leverage all data from an RCT. 
First, we theoretically quantify the estimation bias that arises from naïvely aggregating performance estimates from treatment and control groups and derive the condition under which this bias leads to incorrect model selection. 
Leveraging these theoretical insights, we propose nuisance parameter weighting (NPW), an unbiased model evaluation approach that reweights data from the treatment group to mimic the distribution of samples that would or would not experience the outcome under no intervention.
Using synthetic and real-world datasets, we demonstrate that our proposed evaluation approach consistently yields better model selection than the standard approach, which ignores data from the treatment group, across various intervention effect and sample size settings. 
Our contribution represents a meaningful step towards more efficient model evaluation in real-world contexts.
\end{abstract}

\begin{links}
    \link{Code}{https://github.com/MLD3/NPW}
    \link{Extended version}{https://arxiv.org/abs/2511.05805}
\end{links}

\begin{figure*}[!t]
  \centering
  \includegraphics[width=1.0\linewidth]{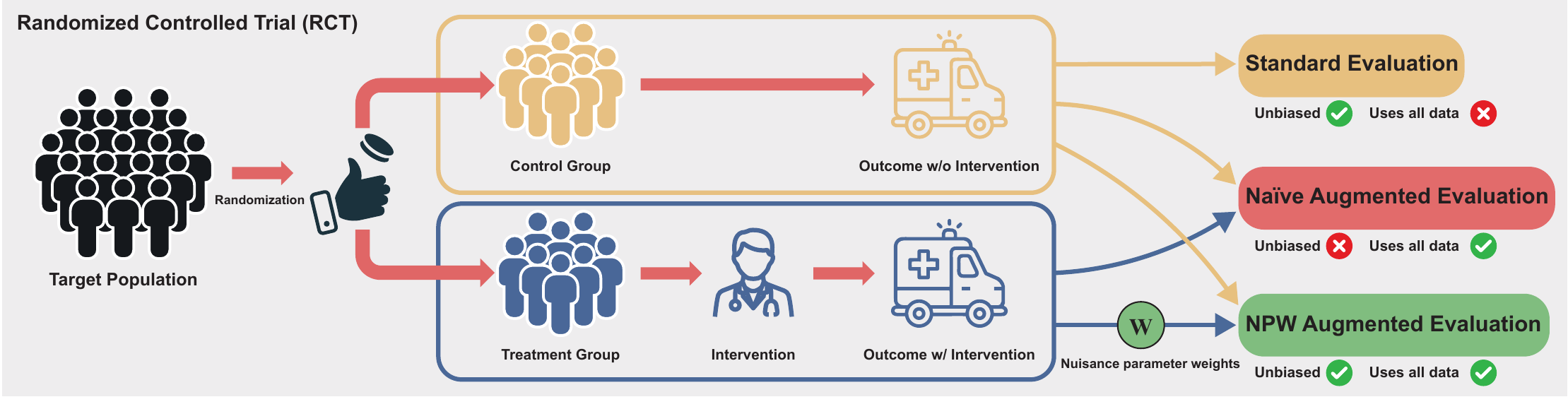}
  \captionof{figure}{Overview of an RCT and different model evaluation approaches using RCT data.
  The standard evaluation is unbiased but only uses data from the control group. 
  \text{Na\"ive} Augmented Evaluation uses data from both the control and treatment groups but is biased.
  Our proposed nuisance parameter weighting (NPW) augmented Evaluation is unbiased and uses all RCT data.}
  \label{fig:overview}
\end{figure*}

\section{Introduction}\label{sec:intro}
Assessing a model's ability to predict the outcome of interest is critical before real-world deployment.
However, in many applications studied by AI for social impact, the presence of an intervention that influences the outcome of interest can complicate model evaluation.
For example, in healthcare, AI models have been developed to predict readmission~\citep{yu2015predicting, huang2021application}.
At the same time, interventions designed to prevent readmission (\textit{e.g.,} post-discharge phone check-in) are used to reduce the overall readmission rate~\citep{hansen2011interventions}.
To allocate limited intervention resources, hospitals often use deterministic rules based on patients' risk index (e.g., the LACE index $>$ 10) to decide who should receive the intervention~\citep{teh2018identifying}.
Similar examples also exist in domains such as:
(1) public health ~\citep{ayer2019prioritizing, deo2015planning, amarasingham2013allocating, wang2023scalable, yang2020modified},
(2) infrastructure maintenance~\citep{gerum2019data, yeter2020risk}, and
(3) education support programs~\citep{mac2019efficacy, adnan2021predicting}.

When measuring model performance in the presence of an intervention, using all data in the evaluation may introduce \textit{outcome bias}~\citep{pajouheshnia2017accounting}, because the intervention can alter the observed outcomes.
Evaluating with un-intervened data (i.e., data in which the intervention was not applied) avoids outcome bias but may introduce \textit{selection bias}, because interventions are often not assigned randomly.
Inverse propensity weighting (IPW)-based methods that reweight data based on their likelihood of intervention can mitigate selection bias, but they require some randomness within the intervention assignment (e.g., based on probabilities)~\citep{pajouheshnia2017accounting, coston2020counterfactual, boyer2023assessing, keogh2024prediction}.
In real-world applications, interventions are often assigned deterministically (\text{e.g.,} based on thresholds), rendering IPW inapplicable.

Temporarily pausing the intervention for a period allows for unbiased model evaluation. 
Though conceptually simple, this approach can be challenging to operationalize in practice. 
Well-established interventions are often supported by a designated team.
A temporary pause may require furloughing the team or appropriately redirecting their efforts. 
Instead, when feasible, a randomized controlled trial (RCT) could eliminate selection bias while maintaining the same level of intervention and, thus, the same personnel effort. 
Although RCTs are often conducted to estimate treatment effects, they also provide unbiased `control' data critical in evaluating model performance in many applications where interventions are otherwise routinely delivered.

During an RCT, the study cohort is randomly assigned to the treatment or control group, and only those in the treatment group receive the intervention.
As a result, both treatment and control groups are representative samples of the study cohort. 
Thus, evaluation using data from the control group yields an unbiased estimate of model performance.

Conducting an RCT is often complex and expensive ~\citep{moore2020variation}.
Therefore, when such data are available, they should be fully leveraged.
However, standard approaches to evaluating a predictive model with these data rely only on the control group, since the goal is to understand the predictive performance of a model in the absence of the intervention.
This restriction reduces the effective sample size, increases variance in performance estimates, and potentially leads to inconclusive results.

In this work, we study how to augment standard evaluation with data from the treatment group in an RCT.
We focus on the area under the receiver operating characteristic curve (AUROC), due to its wide adoption in AI model development and broad applicability to real-world applications~\citep{ling2003auc}.
However, our method can be generalized to any binary classification metric.

We start by considering \text{na\"ive} augmented AUROC, which aggregates AUROC estimates obtained separately from the treatment and control groups.
We theoretically quantify its bias and derive an exact condition under which the bias leads to incorrect model selection.
Building on our theoretical insights, we propose nuisance parameter weighting (NPW) augmented AUROC, an unbiased augmentation approach. 
NPW works by reweighting data from the treatment group to approximate the distribution of samples that would or would not experience the outcome under no intervention.
We empirically show that NPW consistently leads to more accurate model selection across various intervention effect and sample size settings compared to the standard approach.

In summary, our contributions are:
\begin{itemize}
    \item To our knowledge, we are the first to study measuring AUROC using data in which an intervention is applied,
    \item we derive the bias of \text{na\"ive} augmented AUROC and characterize conditions in which it selects incorrect models,
    \item we propose NPW, an unbiased approach to evaluate models using data from both control and treatment groups.
    \item we empirically demonstrate the advantages of our proposed approach in improving model selection.
\end{itemize}

\section{Background}
\subsection{Problem Statement}
Consider a target cohort characterized by $X\in \mathbb{R}^p$.
Each individual is randomly assigned to the treatment ($T=1$) or control ($T=0$) group with probability $\pi$: $T\sim \text{Bern}(\pi)$.
Each individual may experience a binary outcome, $Y$, sampled from a distribution conditioned on their $T$ and $X$:
\begin{equation}
\begin{aligned}
    Y \mid X, T \sim \text{Bern}\!\left(\omega(X) + T\tau(X)\right)
\end{aligned}
\label{eq:outcome_dgp}
\end{equation}
where $\text{Bern}(\cdot)$ is the Bernoulli distribution, 
$\omega(\cdot)$ is the baseline outcome probability, and 
$\tau\!:\!X\!\!\to\!\!(-\omega(X),\,1\!-\!\omega(X))$ represents the treatment effect conditional on $X$. 
The range of $\tau$ ensures that $\mathbf{P}(Y \mid X, T) \in (0,1)$.

Given a dataset $\mathbb{D} = \{(x_i, y_i, t_i)\}_{1\leq i \leq n}$ sampled from the above data generation process (DGP), we aim to evaluate a prediction model, $f\!:\!X\!\!\rightarrow\!\mathbb{R}$, in its AUROC for predicting the outcome under no-intervention, $\mathbb{E}\left[\text{AUC}(\mathbb{X}_0^+, \mathbb{X}_0^-, f)\right]$.
The expectation is over $\mathbb{X}_0^+$ and $\mathbb{X}_0^-$, control samples that did or did not experience the outcome. 
They are defined as follows:
\begin{align*}
 \mathbb{X}_0^+ = \{x_i | t_i=0, y_i=1\}, \quad
 \mathbb{X}_0^- = \{x_i | t_i=0, y_i=0\}.
\end{align*}
The $\text{AUC}(\cdot, \cdot, \cdot)$ notation is defined as:
\begin{align*}
    \text{AUC}(\mathbb{X}^+, \mathbb{X}^-, f)
= \frac{1}{|\mathbb{X}^+||\mathbb{X}^-|}
\sum_{\substack{x^{(i)} \in \mathbb{X}^+ \\ x^{(j)} \in \mathbb{X}^-}}
\mathds{1}_{\left\{ f\!\left(x^{(i)}\right) > f\!\left(x^{(j)}\right) \right\}},
\end{align*}
where $\mathds{1}_{\left\{\cdot\right\}}$ is the indicator function, $\mathbb{X}^+$ and $\mathbb{X}^-$ both denote samples used to compute AUROC, but samples in $\mathbb{X}^+$ have experienced the outcome while samples in $\mathbb{X}^-$ have not.

\subsection{Preliminaries}
To estimate model $f$'s AUROC under no intervention, simply using all data in $\mathbb{D}$ leads to biased estimates, because it produces a weighted average between the target AUROC ($\mathbb{E}\left[\text{AUC}(\mathbb{X}_0^+, \mathbb{X}_0^-, f)\right]$) and three other variant AUROCs.

To see this, we first define the following sets of data based on intervention and outcome variables:
\begin{align*}
    \begin{split}
        \mathbb{D}_0 &= \left\{(x_i, y_i) | t_i=0\right\} \\
        \mathbb{X}_0^- &= \left\{x_i | t_i=0, y_i=0\right\} \\
        \mathbb{X}_0^+ &= \left\{x_i | t_i=0, y_i=1\right\}
    \end{split} 
    \begin{split}
        \mathbb{D}_1 &= \left\{(x_i, y_i) | t_i=1\right\} \\
        \mathbb{X}_1^- &= \left\{x_i | t_i=1, y_i=0\right\} \\
        \mathbb{X}_1^+ &= \left\{x_i | t_i=1, y_i=1\right\}
    \end{split}
\end{align*}

Then, we define short-hand notations for AUROCs estimated with only $\mathbb{D}_0$ and $\mathbb{D}_1$, respectively:
\begin{align*}
    \begin{split}
        \text{AUC}_{\mathbb{D}_0}(f) &= \text{AUC}(\mathbb{X}_0^+, \mathbb{X}_0^-, f)
    \end{split} \\
    \begin{split}
        \text{AUC}_{\mathbb{D}_1}(f) &= \text{AUC}(\mathbb{X}_1^+, \mathbb{X}_1^-, f)
    \end{split}
\end{align*}

Leveraging this notation, the expected AUROC using all data can be expressed as:
\begin{align*}
\mathbb{E}&\left[\text{AUC}_{\text{all}}(f)\right] = \mathbb{E}\left[(1-\pi)^2\underbrace{\text{AUC}_{\mathbb{D}_0}(f)}_{\text{unbiased term}} + \pi^2\underbrace{\text{AUC}_{\mathbb{D}_1}(f)}_{\text{biased term 1}} \right. \\
&\left. \quad + (\pi-\pi^2)\underbrace{\left(\text{AUC}(\mathbb{X}_0^+, \mathbb{X}_1^-, f) + \text{AUC}(\mathbb{X}_1^+, \mathbb{X}_0^-, f)\right)}_{\text{biased term 2}}\right]
\end{align*}
Intuitively, the first term in the expression is unbiased, while the other terms introduce bias.
We refer readers to Appendix Section~\ref{sec:appendix:all-expr} in the extended version of this paper for a detailed derivation of the above expression.

To obtain unbiased estimates of model $f$'s AUROC, the standard approach only uses $\mathbb{D}_0$: $\text{AUC}_{\text{std}}(f) = \text{AUC}_{\mathbb{D}_0}(f)$. 
However, it ignores $\mathbb{D}_1$ entirely, and as a result, the estimate may be associated with high variance when the sample size is small. 
To augment standard AUROC estimates with $\mathbb{D}_1$, a \text{na\"{i}}ve approach averages $\text{AUC}_{\mathbb{D}_0}$ and $\text{AUC}_{\mathbb{D}_1}$ with the randomization probability, $\pi$:
\begin{equation}
\begin{aligned}
    \text{AUC}_{\text{na\"{i}ve}}(f) = (1-\pi)\text{AUC}_{\mathbb{D}_0}(f) + \pi\text{AUC}_{\mathbb{D}_1}(f)
\end{aligned}
\label{eq:naive-auroc}
\end{equation}
The \text{na\"ive} augmented AUROC is biased due to the presence of $\text{AUC}_{\mathbb{D}_1}(f)$. 
However, compared to $\text{AUC}_{\text{all}}(f)$, it has only one bias term, allowing our theoretical analysis to succinctly quantify its bias and characterize scenarios where it incorrectly selects the model as the ground truth AUROC. 

\section{Theoretical Analysis of \text{Na\"ive} Augmentation}

\text{Na\"ively} augmenting the standard AUROC with data from the treatment group leads to biased estimates because the presence of intervention alters the outcomes. 
In this section, we formally derive its theoretical bias.

\subsection{Bias of \text{Na\"ive} Augmented AUROC}
\begin{theorem}[Bias of \text{Na\"ive} Augmented AUROC (Eq.~\ref{eq:naive-auroc})]
\label{thrm:naive-bias}
Let $\mu_0$ and $\mu_1$ be the expected outcome for the control and treatment group, and $\tau$ be the average treatment effect (ATE):
\begin{align*}
    \mu_0 &= \mathbb{E}_X[Y\mid X, T=0] = \mathbb{E}_X[\omega(X)] \\
    \mu_1 &= \mathbb{E}_X[Y\mid X, T=1] = \mathbb{E}_X[\omega(X)+\tau(X)] \\
    \tau &= \mathbb{E}_X[\tau(X)] = \mu_1 - \mu_0
\end{align*}
Under our assumed DGP, the bias of \( \text{AUC}_{\text{na\"ive}}(f) \) is:
\begin{equation}
\begin{aligned}
\text{Bias}\left(\text{AUC}_{\text{na\"ive}}(f)\right) = \alpha\delta(f) - \beta\sigma(f)
\end{aligned}
\label{eq:naive-aug-bias}
\end{equation}
\end{theorem}
where $\alpha$ and $\beta$ are bounded problem-specific real numbers defined as follows:
\begin{align*}
    \alpha = \frac{\pi\tau(1-\mu_0-\mu_1)}{\mu_1(1-\mu_1)} < 1 \quad
    \beta = \frac{\pi}{\mu_1(1-\mu_1)} > 0
\end{align*}
$\delta(f)$ represents model $f$'s true AUROC improvement over a random prediction with AUROC being 0.5. 
$\sigma(f)$ denotes the covariance between the cumulative distribution function (CDF) of model $f$'s prediction and the true individual-level effect, $\tau(X)$, also known as the conditional average treatment effect (CATE), of the intervention.

At a high level, our proof rewrites the bias definition of \text{na\"ive} augmented AUROC using our assumed DGP and Bayes' rule to obtain the simplified bias term in Eq.~\ref{eq:naive-aug-bias}.
Because it does not leverage AUROC-specific techniques, it is generalizable to any binary classification metric.
Detailed derivation of Theorem~\ref{thrm:naive-bias} is provided in Appendix Section~\ref{sec:appendix:naive-avg-bias}.

Theorem~\ref{thrm:naive-bias} shows that the bias of \text{na\"ive} augmented AUROC depends on a linear combination of: (1) the model's true AUROC: $\delta(f)$ and (2) the correlation between model and the CATE function: $\sigma(f)$.
The exact contribution of each factor depends on problem-specific parameters: $\alpha$ and $\beta$.

\subsection{Model Selection with \text{Na\"ive} Augmented AUROC}
To contextualize the potential harm of this bias and better understand how bias in model evaluation could affect downstream decisions in real-world applications, we consider model selection using the estimated AUROC. We derive an exact condition under which model selection using \text{na\"ive} augmented AUROC leads to an incorrect model.
\begin{theorem}[Condition for Incorrect Model Selection with \text{Na\"ive} Augmented AUROC]
\label{thrm:naive-select}
Given models $f_1$ and $f_2$, let $\theta(f)$ and $\hat{\theta}(\cdot)$ be models' true AUROC and expected AUROC estimate using \text{na\"ive} augmented evaluation, respectively. 
Without loss of generality, we consider $\hat{\theta}(f_1) > \hat{\theta}(f_2)$. 
When $\hat{\theta}(f_1)-\hat{\theta}(f_2) > \beta(\delta(f_1)-\delta(f_2))$, we will always have $\theta(f_1) < \theta(f_2)$, therefore $\hat{\theta}(\cdot)$ selects the incorrect model.
\end{theorem}
\proof
    Our proof starts by expressing the expected AUROC estimate from \text{na\"ive} augmentation in terms of its bias:
    \begin{align*}
        \hat{\theta}(f) = \theta(f) - \text{Bias}\left(\hat{\theta}(f)\right)
    \end{align*}
Applying Theorem~\ref{thrm:naive-bias} and the definition of $\delta(f)$, we get:
    \begin{align*}
        \hat{\theta}(f) = &\left(1-\alpha\right)\theta(f) + \alpha0.5 + \beta\sigma(f)
    \end{align*}
Given the above rewrite of $\hat{\theta}(f_1)$, we express the difference between $\theta(f_1)$ and $\theta(f_2)$ as: 
\begin{align*}
    \theta(f_1)-\theta(f_2) = \frac{(\hat{\theta}(f_1) - \hat{\theta}(f_2)) - \beta(\sigma(f_1) - \sigma(f_2))}{1-\alpha}
\end{align*}
Because by definition $\alpha < 1$, the denominator of the right-hand side is positive. 
$\theta(f_1) < \theta(f_2)$ holds when
\begin{align*}
    \hat{\theta}(f_1) - \hat{\theta}(f_2) < \beta\left(\sigma(f_1) - \sigma(f_2)\right)
\end{align*}

Intuitively, we can think of this condition as two scenarios. In the first scenario, if the model selected by the \text{naïve} augmented AUROC has a lower or equal correlation with CATE than the model not selected, then \text{naïve} augmented AUROC will \textit{never} choose the wrong model.
This is because the right-hand side of the condition remains non-positive, while the left-hand side is always positive.
In the second scenario, the model selected by the \text{naïve} augmented AUROC correlates more with CATE than the model not selected. 
The \text{na\"ive} augmented AUROC selects the incorrect model if the 
estimated AUROC difference between the two models is smaller than a scaled version of their difference in CATE correlation.
The scaling factor, $\beta$, is problem-specific but is positive and unbounded. 
Therefore, a large $\beta$ in some scenarios can make the incorrect selection unavoidable as the estimated AUROC difference is upper bounded by 1.

\section{Unbiased Augmentation with NPW}
Here, we propose \textbf{N}uisance \textbf{P}arameter \textbf{W}eighting (NPW) augmented AUROC that leads to unbiased estimates while leveraging data from both control and treatment groups.

Similar to \text{na\"ive} augmentation, NPW estimates AUROC by averaging the standard estimate with an alternative estimate produced based on data from the treatment group, $\mathbb{D}_1$:
\begin{align*}
    \text{AUC}_{\text{NPW}}(f) = (1-\pi)\text{AUC}_{\mathbb{D}_0}(f) + \pi\text{AUC}_{\text{alt}}(f)
\end{align*}
The key to produce an unbiased $\text{AUC}_{\text{alt}}(f)$ lies in recovering distributions of samples that would or would not experience the outcome \textit{without} intervention, denoted by  $\mathbf{P}(X_0^+)$ and $\mathbf{P}(X_0^-)$.
Because $\mathbb{D}_1$ is sampled from distributions of data that would or would not experience the outcome \textit{with} intervention, denoted by  $\mathbf{P}(X_1^+)$ and $\mathbf{P}(X_1^-)$, \text{na\"ively} computing $\text{AUC}_{\text{alt}}(f)$ with $\mathbb{D}_1$ leads to biased estimates.

Our proposed NPW uses two weighting approaches to recover $\mathbf{P}(X_0^+)$ and $\mathbf{P}(X_0^-)$.
The first approach uses data from the control group, $\mathbb{D}_0$, to learn a model that estimates the probability of a sample experiencing the outcome without intervention. 
Then, it re-weights $\mathbb{D}_1$ based on this probability to recover $\mathbf{P}(X_0^+)$ and $\mathbf{P}(X_0^-)$.
However, this approach ignores the observed outcomes in $\mathbb{D}_1$, which provides information about $\mathbf{P}(X_1^+)$ and $\mathbf{P}(X_1^-)$.
To leverage this information, we design a second approach to recover $\mathbf{P}(X_0^-)$ and $\mathbf{P}(X_0^+)$ by correcting for the effect or intervention in $\mathbf{P}(X_1^-)$ and $\mathbf{P}(X_1^+)$ using estimates of CATE.

The following section provides high-level derivations of the two weighting approaches.
Additional details are available in Appendix Section~\ref{sec:appendix:npw}.

\subsection{Derivations of NPW}
To derive the \textbf{first approach}, we write $\mathbf{P}(X_0^-)$ and $\mathbf{P}(X_0^+)$ using the Bayes' rule and the assumed DGP, leading to the following formulations in terms of weighted $\mathbf{P}(X)$, the natural distribution of samples in the target cohort:
\begin{equation}
\begin{aligned}
    \mathbf{P}(X_0^-) =  \frac{1-\omega(X)}{1-\mu_0}\mathbf{P}(X),\  \mathbf{P}(X_0^+) = \frac{\omega(X)}{\mu_0}\mathbf{P}(X)
\end{aligned}
\label{eq:first_approach}
\end{equation}
where the weights are based on  $\omega(X)$ and $\mu_0$, the individual and average probability of experiencing the outcome without intervention.
Equation~\ref{eq:first_approach} shows that we can recover $\mathbf{P}(X_0^+)$ or $\mathbf{P}(X_0^-)$ by up-weighting samples with higher or lower probability of experiencing the outcome, receptively.

Because the intervention is random in RCT, $\mathbb{D}_1$ contains random samples from the target cohort.
Therefore, we can approximate $\mathbf{P}(X_0^-)$ and $\mathbf{P}(X_0^+)$ by re-weighting $\mathbb{D}_1$ with $\hat{\omega}(X)$, an unbiased estimate of $\omega(X)$, following Equation~\ref{eq:first_approach}.
Formally we denote the AUROC estimated with this approach as: $\text{AUC}_{\hat{\omega}}(f) = \text{AUC}(\mathbb{X}_{\hat{\omega}}^+, \mathbb{X}_{\hat{\omega}}^-, f)$, where $\mathbb{X}_{\hat{\omega}}^+$ and $\mathbb{X}_{\hat{\omega}}^-$ are $\mathbb{D}_1$ reweighted to approximate $\mathbf{P}(X_0^+)$ and $\mathbf{P}(X_0^-)$.

To derive the \textbf{second approach}, we similarly apply the Bayes' rule and the DGP to rewrite $\mathbf{P}(X_0^-)$ and $\mathbf{P}(X_0^+)$ in terms of $\mathbf{P}(X_1^-)$ and $\mathbf{P}(X_1^+)$ as follows:
\begin{equation}
    \begin{aligned}
    \mathbf{P}(X_0^-) &=\frac{1-\mu_1}{1-\mu_0}\mathbf{P}(X_1^-) + \frac{\tau(X)}{1-\mu_0}\mathbf{P}(X) \\
    \mathbf{P}(X_0^+) &=\frac{\mu_1}{\mu_0}\mathbf{P}(X_1^+) - \frac{\tau(X)}{\mu_0}\mathbf{P}(X)
\end{aligned}
\label{eq:second_approach}
\end{equation}
This formulation leads to AUROC estimates with $\hat{\tau}(X)$, an unbiased estimate of $\tau(X)$: $\text{AUC}_{\hat{\tau}}(f) = \text{AUC}(\mathbb{X}_{\hat{\tau}}^+, \mathbb{X}_{\hat{\tau}}^-, f)$, where $\mathbb{X}_{\hat{\tau}}^+$ and $\mathbb{X}_{\hat{\tau}}^-$ are $\mathbb{D}_1$ reweighted following Equation~\ref{eq:second_approach}.

The second approach recovers $\mathbf{P}(X_0^-)$ by up-weighting samples in $\mathbb{D}_1$ with higher CATE, because they are more likely not to experience the outcome without intervention.
Conversely, to recover $\mathbf{P}(X_0^+)$, it down-weights higher CATE samples in $\mathbb{D}_1$ to account for the possibility that they experienced the outcome primarily due to the intervention. 

Since $\text{AUC}_{\hat{\omega}}(f)$ and $\text{AUC}_{\hat{\tau}}(f)$ each depend on a nuisance parameter estimate that could have high variance in practice, we average them to further reduce the estimation variance.
The \textbf{final form} of the alternative estimate in NPW is:
\begin{align*}
    \text{AUC}_{\text{alt}}(f) = \frac{\text{AUC}_{\hat{\omega}}(f) + \text{AUC}_{\hat{\tau}}(f)}{2}
\end{align*}

\subsection{Practical Considerations}
Calculating $\text{AUC}_{\hat{\omega}}(f)$ and $\text{AUC}_{\hat{\tau}}(f)$ with weighted sampling can be inefficient, as $\text{AUC}_{\hat{\tau}}(f)$ involves aggregating two distributions through their samples.
Therefore, we implement them with weighted AUROC, which is equivalent to weighted sampling. 
See Appendix Section~\ref{sec:appendix:npw} for the exact formulation of weighted AUROC.

\section{Experimental Setup}
Does NPW empirically improve model evaluation and, in turn, model selection?
In this section, we describe our experimental setup for investigating this question. 

\subsection{Datasets}
We design a synthetic dataset to validate our theoretical results and evaluate NPW under controlled settings. 
Leveraging two real-world datasets, we evaluate NPW in realistic settings, where the nuisance parameters must be estimated.

The \textbf{synthetic} dataset is generated as follows:
\begin{align*}
    &x_i \sim \mathcal{N}(0,\,\mathbf{I}_{20}), \quad t_i \sim \text{Bern}(0.5), \\
    &\tau_i \mid x_i = \frac{\sigma(w_\tau x_i)\big(1 - w_y x_i\big)\Delta}
        {\frac{1}{n}\sum_{j=1}^n \sigma(w_\tau x_j)\big(1 - w_y x_j\big)}, \\
    &y_i \mid t_i, x_i \sim \text{Bern}\!\left(\sigma(w_y x_i) + t_i \tau_i\right),
\end{align*}
where $w_y \in \mathbb{R}^{20}$ has $40\%$ of entries drawn from $\mathcal{N}(0,1)$ and the rest set to $0$, 
and $w_\tau \in \mathbb{R}^{20}$ is sampled from $\{0, 0.1, 0.2, 0.3, 0.4\}$ with probabilities $\{0.8, 0.05, 0.05, 0.05, 0.05\}$. 
The scalar $\Delta \in [-1, 1]$ controls the average treatment effect (ATE). 
We generate $100{,}000$ samples in total, using the AUROC computed on all control samples as the ground truth, 
and subsample $n=200$ points to mimic the limited size of an RCT.

We use \textbf{AMR-UTI} dataset ($n=15,806$) to simulate an RCT with real-world data.
It is a dataset containing antimicrobial resistance results for patients with urinary tract infections (UTIs) ~\cite{oberst2020amr}.
AMR-UTI covers four types of antimicrobial treatments and uniquely provides ground-truth antimicrobial resistance labels for each treatment obtained via microbiology testing.
We transform it into an RCT dataset by considering the two most common treatments: nitrofurantoin (NIT) and trimethoprim-sulfamethoxazole (SXT) as the treatment and control, respectively.
Patients are randomly assigned to the treatment or control group with a probability of $0.5$. 
The antimicrobial resistance labels are considered the outcome of interest. 
Since each pathogen associated with patient infection is tested for resistance to each antimicrobial, we have ground truth outcomes under both treatment and control settings.
In our resampled RCT dataset, we observe the NIT resistance label for patients in the treatment group and the SXT resistance label for patients in the control group.

Lastly, we leverage the \textbf{Readmission} dataset, a real-world RCT dataset ($n=1,518$), collected from Michigan Medicine (MM), a large academic medical center associated with the University of Michigan, to evaluate our approach with real RCT data.
In this dataset, discharged patients are randomly assigned to the treatment or control group.
Patients in the treatment group receive post-discharge phone check-in calls as an intervention. 
The outcome of interest is whether a patient experiences an unplanned readmission to the MM within 30 calendar days of discharge.
Given that the study randomizes an existing intervention already in use at MM, the University of Michigan institutional review board (IRB) determined this RCT was a quality improvement (QI) initiative exempt from human subjects research, waiving the need for patient consent.
One of the goals of this RCT is to determine whether the hospital should adopt a newly developed Epic readmission risk prediction model~\citep{hwang2021external} in its risk stratification workflow.
The decision primarily depends on whether the Epic model significantly outperforms the LACE index~\citep{van2010derivation}, a model currently used by the hospital to estimate readmission risk.
Given the difficulty of predicting readmission, we expect only modest performance gains from the Epic model. 
Thus, standard evaluation could require a large sample size for a significant comparison.
Our experiments evaluate different AUROC estimates using a statistical hypothesis testing framework, measuring whether they can reject the null hypothesis of Epic \textit{does not} improve over LACE with limited samples.
We provide additional details about this dataset in Appendix Section~\ref{appendix:readmission}.

\subsection{Models}

To mimic a model selection scenario, we needed to generate several models to select based on their AUROC estimates.
For the Synthetic and AMR-UTI datasets, we use gradient-boosted decision trees~\citep{friedman2001greedy} to create models of varying performance. We trained each model with a varying number of samples, ranging from $100$ to $1500$ (for synthetic), and from $10$ to $5,806$ (for AMR-UTI).
To ensure a uniform distribution of models' true AUROC, we randomly sample one model from each $0.005$ increment in true AUROC.
This sampling procedure results in $65$ models in the synthetic dataset and $31$ in the AMR-UTI dataset.

For the Readmission dataset, we use the prospective predictions from the Epic model and LACE index during the RCT for evaluation; therefore, only existing models are applied, and no model training is necessary.

\begin{figure*}[!t]
  \centering
  \includegraphics[width=1.0\linewidth]{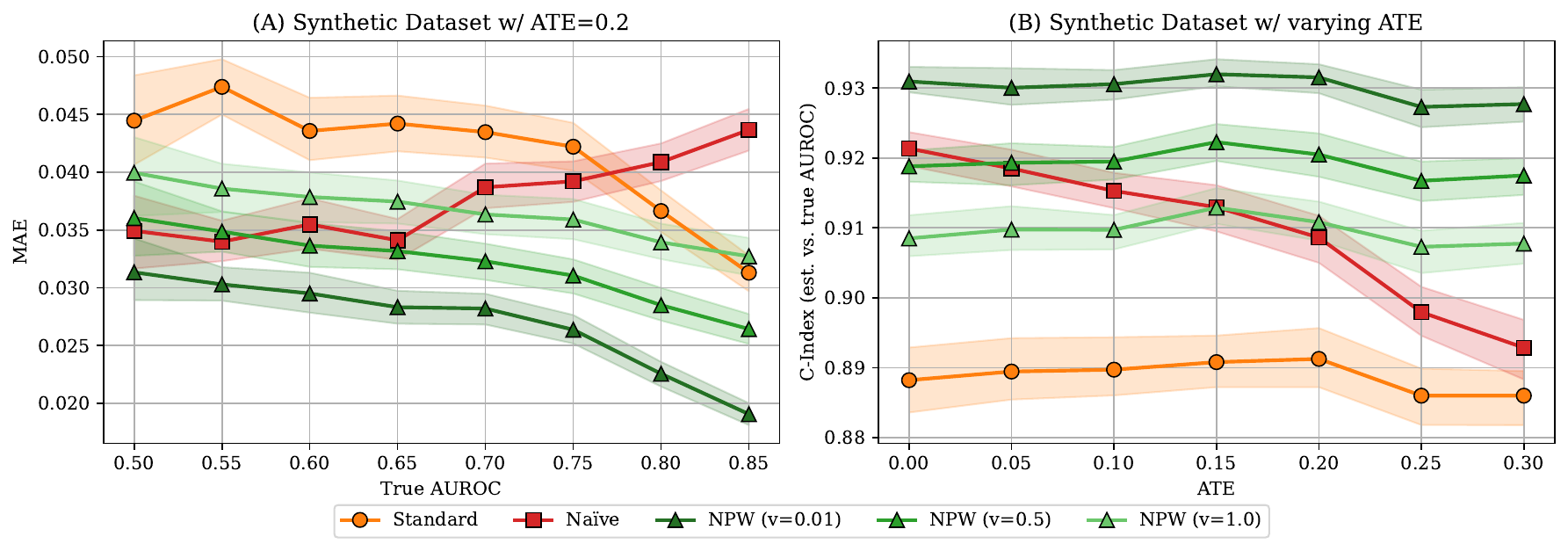}
  \captionof{figure}{Empirical results with the synthetic dataset.
  (A) MAEs of different AUROC estimates for models with various ground true AUROCs. 
  (B) C-index of model rankings induced by different AUROC estimates evaluated under interventions of various ATEs.
  In both figures, NPW consistently outperforms the standard approach, and its advantage over the standard approach increases as the variance ($v$) of nuisance parameter estimates decreases. 
  Error bars are bootstrapped 95\% confidence intervals.}
  \label{fig:sim}
\end{figure*}

\subsection{Nuisance Parameter Estimation}
Our proposed NPW augmented AUROC requires estimating nuisance parameters, $\omega(X)$ and $\tau(X)$, to produce AUROC estimates.
The nuisance parameters are directly available in the synthetic dataset due to the known DGP. 
Therefore, we simulate unbiased nuisance parameter estimates by adding Gaussian noise with variance $v$ to the true nuisance parameters.
We estimate nuisance parameters with the cross-fitting approach for the AMR-UTI and Readmission datasets.
Specifically, when evaluating models with $n$ data points, the nuisance parameters are also estimated with the same $n$ data points.
Data are split into $k$ folds. 
For each fold, we use the remaining $k-1$ folds to fit a nuisance parameter model with gradient-boosted decision trees, which then predicts nuisance parameters for the current fold.

\subsection{Evaluation Metrics}
In the synthetic dataset, we evaluate the quality of the AUROC estimate using the mean absolute error (MAE).
For the AMR-UTI dataset, we evaluate whether AUROC-based model rankings align with the true rankings using the concordance index (C-index), which quantifies their agreement.

In the readmission dataset, we assess AUROC estimates by their statistical power to detect a true performance improvement between Epic and LACE models.
Power is defined as the proportion of bootstrap-derived P-values below the significance level $\alpha=0.05$.
Each P-value is estimated from $1{,}000$ bootstrap samples by comparing the AUROCs of LACE and Epic and calculating the proportion of samples where LACE outperforms Epic.
We report mean values with 95\% confidence intervals to summarize each metric.

\subsection{Baselines}
We evaluate two baselines to contextualize the performance of our proposed approach.
(1) \textbf{Standard approach}: estimating AUROC with RCT data from the control group only.
(2) \textbf{\text{Na\"ive} approach}: \text{na\"ively} averaging AUROC estimated with RCT data from both the control and treatment groups.

\section{Empirical Results}
We evaluate NPW augmented AUROC in terms of three aspects compared to the baseline approaches:
(1) Does NPW reduce estimation error?
(2) Does NPW improve model selection performance?
(3) Does NPW boost the statistical power of hypothesis testing in a real-world RCT?
\subsection{NPW Augmentation Reduces Estimation Error}
We start by evaluating the NPW and baselines in terms of their MAE.
We use a synthetic dataset with ATE = $0.2$ to compare all approaches in evaluating models with various true AUROCs.
Because NPW's performance depends on the nuisance parameter estimates, we also simulate estimates with varying qualities by controlling their variances ($v$), with lower $v$ indicating higher-quality estimates.

As shown in Figure~\ref{fig:sim}(A), when the quality of nuisance parameter estimates is high ($v=0.01$), NPW has lower average MAE than both the standard and \text{na\"ive} approaches across all evaluated models. 
As $v$ increases, NPW's MAE worsens due to increased variance in the estimation.
However, when the quality of nuisance parameter estimates is poor ($s=1.0$), NPW still outperforms the standard approach across most models and beats the \text{na\"ive} approach on models with higher true AUROC (\text{i.e.}, true AUROC $\ge$ 0.7).

Because NPW augmented AUROC removes the intervention effect on the treated data and is unbiased, we expect it to be consistent with different intervention ATEs. 
Additional results with other ATEs can be found in the Appendix Section~\ref{appendix:exp:sim_ates}.
The results show similar trends to Figure~\ref{fig:sim}(A).

Our results also show that \text{na\"ive} approach only reduces MAE compared to the standard approach when the intervention's ATE or model's true AUROC is low, and hurts the performance when either quantity is high, confirming our theory and highlighting the importance of unbiased augmentation with NPW. 
See Appendix Section~\ref{appendix:exp:sim_ates} for details. 

\begin{figure}[!t]
  \centering
  \includegraphics[width=1.0\linewidth]{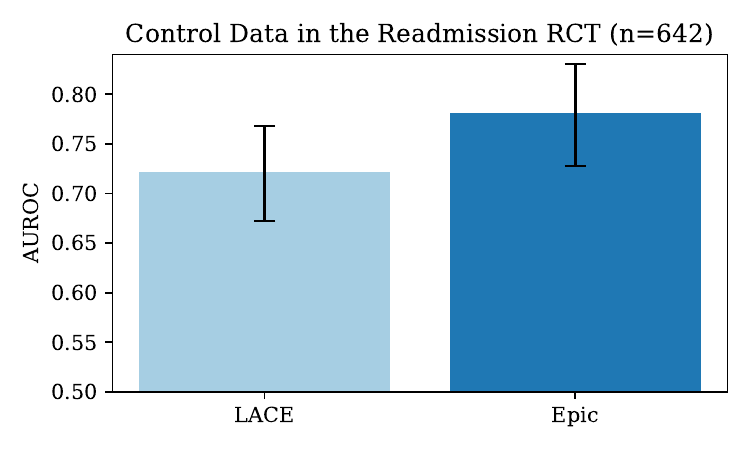}
  \captionof{figure}{LACE and Epic's readmission prediction performance in terms of the AUROC, estimated using all control data. Error bars are bootstrapped 95\% confidence intervals.}
  \label{fig:epic_vs_lace}
\end{figure}

\begin{figure*}[!t]
  \centering
  \includegraphics[width=1.0\linewidth]{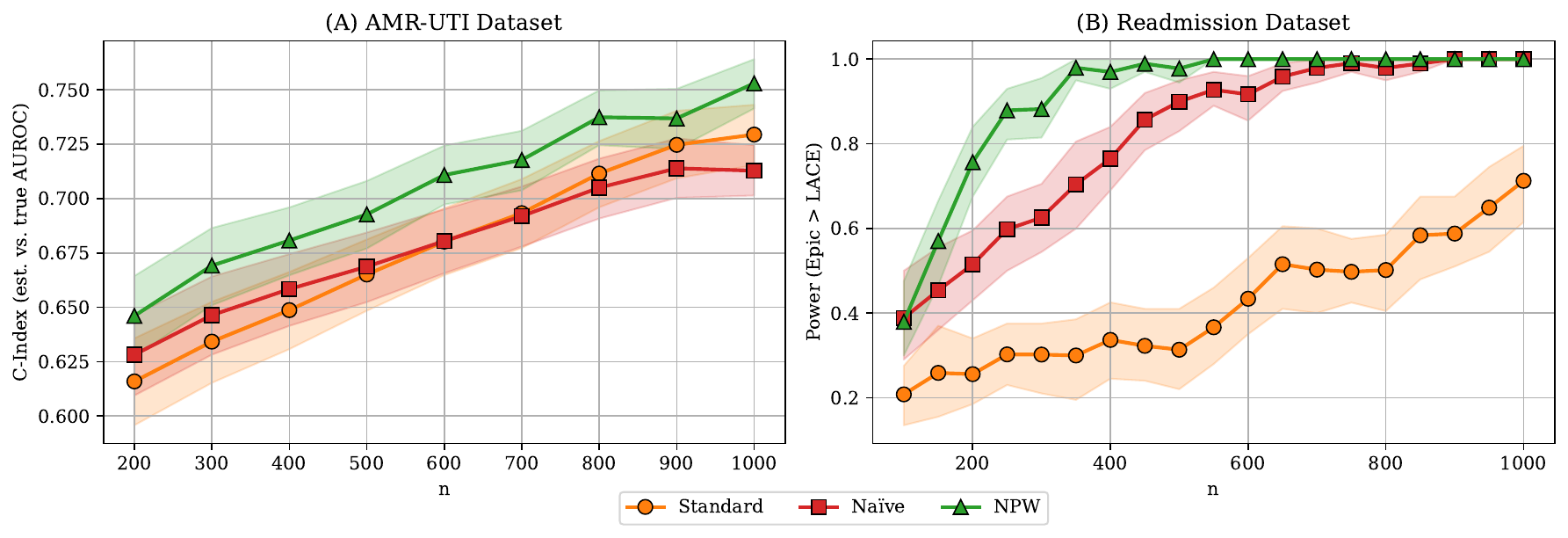}
  \captionof{figure}{
  Empirical results with the real-world datasets.
  (A): C-index of model performance rankings induced by different AUROC estimates on the AMR-UTI dataset.
  (B): Statistical power for testing whether Epic outperforms LACE using different AUROC estimates on the Readmission dataset.
  On both figures, NPW achieves the highest C-index and statistical power across all sample size settings ($n$).
  Error bars are bootstrapped 95\% confidence intervals.}
  \label{fig:real}
\end{figure*}

\subsection{NPW Augmentation Improves Model Ranking}
AUROC estimates are commonly used to rank models and select the best-performing one.  
Do AUROC estimates with lower MAE lead to improved model ranking performance?
Here, we evaluate different AUROC estimation approaches in terms of the C-index of their induced model rankings against those determined by the models' true AUROCs. 
A higher C-index indicates better model ranking performance.

As shown in Figure~\ref{fig:sim}(B), NPW augmented AUROCs consistently improve the C-index compared to the standard AUROC across a wide range of intervention ATEs.
The improvement is robust against degradations in the quality of nuisance parameter estimates. 
Similar to Figure~\ref{fig:sim}(A), NPW's C-index improvement over the standard AUROC diminishes as $v$ increases.
With high-quality nuisance parameter estimates ($v=0.01$), NPW outperforms the \text{na\"ive} approach across all ATEs.
Even with poor nuisance parameter estimates ($v=1.0$), NPW remains advantageous over the \text{na\"ive} approach in settings with higher ATEs (ATE$>0.15$).
This is due to \text{na\"ive} approach's increasing estimation bias as the ATE increases, confirming our theoretical results.

In the AMR-UTI dataset, the nuisance parameters are estimated using the same samples for estimating AUROC.
As shown in Figure~\ref{fig:real}(A), NPW augmented AUROC consistently improves over the standard approach when the sample sizes used to estimate nuisance parameters and estimate AUROC is less than or equal to $1000$.
As expected, increasing the number of samples diminishes the performance gap between NPW and standard AUROC. However, real-world RCT datasets often have limited sample sizes, underscoring the practical benefits of NPW in such scenarios.

Comparing Figures~\ref{fig:sim}(B) and~\ref{fig:real}(A), we see that \text{na\"ive} approach improves model selection in the synthetic dataset, but worsens it in the AMR-UTI dataset. This is because the condition for \text{na\"ive} augmented AUROC to select the wrong model (Theorem~\ref{thrm:naive-select}) is more likely to hold in the AMR-UTI dataset, which supports our theoretical insights. See Appendix Section~\ref{appendix:exp:naive_compare} for more details.

\subsection{NPW Augmentation Boosts Statistical Power}
Can NPW improve the statistical power of hypothesis testing in real-world RCTs? 
We use the Readmission RCT dataset to investigate this question.
We aim to test whether the Epic model outperforms LACE in predicting readmission. 
Testing this hypothesis is challenging because the Epic model only moderately improves over LACE (Figure~\ref{fig:epic_vs_lace}).

We evaluate the statistical power to test this hypothesis using different approaches and sample sizes.
As shown in Figure~\ref{fig:real}(B), all methods yield higher power as the sample size increases, indicating more confidence in Epic's improvement over LACE.
To achieve moderately high power (\textit{i.e.}, 0.8), the standard approach requires over $1000$ samples, of which 423 control samples are used, given the 58\% treatment rate in the RCT. 
Under the same setting, the \text{na\"ive} approach, which uses all the data but is biased, requires close to 500 samples.
In contrast, when using NPW, we obtain the same level of power with only $200$ samples, a five-times improvement in sample efficiency over the standard approach.

\section{Conclusion}
In this work, we investigate model evaluation using AUROC with RCT data.
We propose augmenting the standard evaluation protocol by leveraging treatment data to reduce estimation error and improve downstream model selection performance.
We theoretically quantify the bias of \text{na\"ive} augmentation and derive the exact condition under which it leads to incorrect model selection.
Leveraging these insights, we propose NPW, an unbiased approach that reweights data from the treatment group to mimic the distribution of individuals who would or would not experience the outcome in the absence of intervention. 
As a weighting-based approach, NPW generalizes to any binary classification metric.
We empirically validate NPW and demonstrate that it enhances AUROC estimation, model selection, and hypothesis testing across diverse synthetic and real-world datasets.

Our work is not without limitations. First, we focus on the RCT setting, where interventions are assigned at random. 
Although restrictive, this is often necessary since interventions are frequently assigned deterministically.
Second, implementing NPW requires estimating nuisance parameters. 
Therefore, our empirical results heavily depend on the quality of the nuisance parameter estimates, which can vary between applications. Encouragingly, our approach yields better estimates of model performance in a real-world setting with limited data.
Practitioners applying NPW to their problem should first rigorously assess the quality of their nuisance parameter estimates.
Finally, theoretical insights in this work focus on analyzing the bias associated with the estimates.
Future work should examine questions regarding the variance of augmented AUROCs.

Despite these limitations, researchers evaluating models with RCTs should consider using NPW augmentation. 
This approach theoretically yields unbiased estimates of model performance and can make the most of RCT data, potentially reducing the length and cost of RCT studies.

\section{Acknowledgements}
This work was supported by the University of Michigan's AI \& Digital Health Innovation, the Michigan Institute for Data \& AI in Society's Propelling Original Data Science Grant, and the National Science Foundation (NSF) under Award No. 2124127.
We thank Trenton Chang, Donald Lin, Donna Tjandra, Gregory Kondas, Jung Min Lee, Meera Krishnamoorthy, Michael Ito, Paco Haas, Sarah Jabbou, Stephanie Shepard, and Zhiyi Hu for their helpful conversations and feedback on this work.

\bibliography{references}

\newpage
\onecolumn
\section*{\huge Appendix}
\setcounter{secnumdepth}{2}
\ifdefined\isMain
\else

    \documentclass[letterpaper,11pt]{article}
    \usepackage[margin=1in]{geometry}
    \usepackage{times}
    \usepackage{graphicx}
    \usepackage{titlesec}
    \usepackage{times}
    \usepackage{helvet}
    \usepackage{courier}
    \usepackage[hyphens]{url}
    \usepackage{graphicx}
    \usepackage{natbib}
    \usepackage{caption}
    \usepackage{amsmath}
    \usepackage{amssymb}
    \usepackage{graphicx}
    \usepackage{caption}
    \usepackage{dsfont}
    \usepackage{tikz}
    \usepackage{amsthm}
    \usepackage{subfigure}
    \usepackage{times}
    \usepackage{helvet}
    \usepackage{courier}
    \usepackage{xcolor}
    
    \newcommand{\unif}{\pazocal{U}}
    \newcommand*\circled[1]{\tikz[baseline=(char.base)]{
                \node[shape=circle,draw,inner sep=2pt] (char) {#1};}}
    
    \titleformat{\section}{\large\bfseries}{\thesection.}{1em}{}
    \titleformat{\subsection}{\normalsize\bfseries}{\thesubsection}{1em}{}
    
    \title{{\Large\textbf{Appendix: Measuring Model Performance in the Presence of an Intervention}}}
    \author{\textbf{Anonymous submission}}
    \date{}
    
    \begin{document}

\fi

\maketitle

\section{Bias of \text{Na\"ive} augmented AUROC}
\label{sec:appendix:naive-avg-bias}
From the definition of bias, we have the bias of \text{Na\"ive} augmentation as:
\begin{align*}
    \text{Bias}\left(
    \text{AUC}_{\text{na\"ive}}(f)\right) =
    \pi \bigg(\mathbb{E}\left[\text{AUC}_{\text{std}}(f)\right] - \mathbb{E}\left[\text{AUC}_{\text{igr}}(f)\right]\bigg)
\end{align*}
Applying the definitions of AUROC, we obtain the following:
\begin{align*}
&\mathbb{E}\left[\text{AUC}_{\text{std}}\left(f\right)\right] = \mathbb{E}\left[\text{AUC}\left(\mathbb{X}_0^+, \mathbb{X}_0^-, f\right)\right] = \mathbf{P}\left(f(X^+) > f(X^-)\right), \\
&X^+ \sim \mathbf{P}(X|Y=1, T=0), X^- \sim \mathbf{P}(X|Y=0, T=0) \\
&\mathbb{E}\left[\text{AUC}_{\text{igr}}\left(f\right)\right] = \mathbb{E}\left[\text{AUC}\left(\mathbb{X}_1^+, \mathbb{X}_1^-, f\right)\right] = \mathbf{P}\left(f(X^+) > f(X^-)\right), \\
&X^+ \sim \mathbf{P}(X|Y=1, T=1), X^- \sim \mathbf{P}(X|Y=0, T=1) 
\end{align*}
Rewriting the above probabilities in terms of their integral forms  leads to the following:
\begin{align*}
    \mathbb{E}\left[\text{AUC}_{\text{std}}\left(f\right)\right] = \iint \mathbf{P}(x_i|Y=1, T=0)\mathbf{P}(x_j|Y=0, T=0) \mathds{1}_{\{f(x_i)>f(x_j)\}}dx_idx_j \\
    \mathbb{E}\left[\text{AUC}_{\text{igr}}\left(f\right)\right]  = \iint \mathbf{P}(x_i|Y=1, T=1)\mathbf{P}(x_j|Y=0, T=1) \mathds{1}_{\{f(x_i)>f(x_j)\}} dx_idx_j
\end{align*}
where $\mathds{1}\{\cdot\}$ denotes the indicator function.
Next, we apply the Bayes rule, the assumed DGP, and the assumption that intervention ($T$) is randomly assigned to simplify the conditional probabilities:
\begin{equation}
\begin{aligned}
    \mathbf{P}(X|Y=0, T=0) &= \frac{\mathbf{P}(Y=0 | X, T=0)\mathbf{P}(X | T=0)}{\mathbb{E}_X[\mathbf{P}(Y=0 | X, T=0)]} = \frac{1-\omega(X)}{1-\mathbb{E}_X[\omega(X)]}\mathbf{P}(X) \\
    \mathbf{P}(X|Y=1, T=0) &= \frac{\mathbf{P}(Y=1 | X, T=0)\mathbf{P}(X | T=0)}{\mathbb{E}_X[\mathbf{P}(Y=1 | X, T=0)]} = \frac{\omega(X)}{\mathbb{E}_X[\omega(X)]}\mathbf{P}(X) \\
    \mathbf{P}(X|Y=0, T=1) &= \frac{\mathbf{P}(Y=0 | X, T=1)\mathbf{P}(X | T=1)}{\mathbb{E}_X[\mathbf{P}(Y=0 | X, T=1)]} = \frac{1-(\omega(X)+\tau(X))}{1-\mathbb{E}_X[\omega(X)+\tau(X)]}\mathbf{P}(X) \\
     \mathbf{P}(X|Y=1, T=1) &= \frac{\mathbf{P}(Y=1 | X, T=1)\mathbf{P}(X | T=1)}{\mathbb{E}_X[\mathbf{P}(Y=1 | X, T=1)]} 
    = \frac{\omega(X)+\tau(X)}{\mathbb{E}_X[\omega(X)+\tau(X)]}\mathbf{P}(X) 
\end{aligned}
\label{eq:proba_bayes}
\end{equation}

For notational simplicity, we further define the following population-level constants:
\begin{align*}
    \mu_0 &= \mathbb{E}_X[\mathbf{P}(Y=1 | X, T=0)] = \mathbb{E}_X[\omega(X)]\\
    \mu_1 &= \mathbb{E}_X[\mathbf{P}(Y=1 | X, T=1)] = \mathbb{E}_X[\omega(X)-\tau(X)]\\
    \tau &= \mu_1 - \mu_0 = \mathbb{E}_X[\tau(X)]
\end{align*}

Intuitively, $\mu_0$ and $\mu_1$ represent the population's probability of outcome in the absence of intervention and in the presence of intervention, respectively, and $\tau$ indicates the average treatment effect (ATE) of the intervention. 
Substituting the above constants, we further simplify the conditional probabilities from Equation~\ref{eq:proba_bayes}:
\begin{equation}
    \begin{split}
        \mathbf{P}(X|Y=0, T=0) &=  \frac{1-\omega(X)}{1-\mu_0}\mathbf{P}(X)\\ 
        \mathbf{P}(X|Y=1, T=0) &= \frac{\omega(X)}{\mu_0}\mathbf{P}(X) \\
    \end{split}\qquad
    \begin{split}
        \mathbf{P}(X|Y=0, T=1) &= \frac{1-\omega(X)-\tau(X)}{1-\mu_1}\mathbf{P}(X)\\  
        \mathbf{P}(X|Y=1, T=1) &= \frac{\omega(X)+\tau(X)}{\mu_1}\mathbf{P}(X)
    \end{split}
\label{eq:proba_bayes_simple}
\end{equation}

The above expressions of the conditional probabilities rely on Bayes' rule.
Therefore, they assume that $\mu_0$, $\mu_1$, $1-\mu_0$, and $1-\mu_1$ are all non-zero.
This is a reasonable assumption because these terms are only zero when the outcome is homogeneous (e.g., every individual has the same outcome).
In such a scenario, AUROC is undefined.

Given the above expressions, we can rewrite $\mathbb{E}\left[\text{AUC}_{\text{igr}}\left(f\right)\right]$ as the following:
\begin{align*}
    \mathbb{E}&\left[\text{AUC}_{\text{igr}}\left(f\right)\right] = \iint \frac{\omega(x_i)+\tau(x_i)}{\mu_1}\frac{1-\omega(x_j)-\tau(x_j)}{1-\mu_1}\mathbf{P}(x_i)\mathbf{P}(x_j) \mathds{1}{\{f(x_i)>f(x_j)\}} dx_idx_j \\
    &= \underbrace{\iint \frac{\omega(x_i)(1-\omega(x_j))\mathbf{P}(x_i)\mathbf{P}(x_j)}{\mu_1(1-\mu_1)} \mathds{1}_{\{f(x_i)>f(x_j)\}} dx_idx_j}_{\circled{A}} - \underbrace{\iint \frac{\tau(x_i)\tau(x_j)\mathbf{P}(x_i)\mathbf{P}(x_j)}{\mu_1(1-\mu_1)} \mathds{1}_{\{f(x_i)>f(x_j)\}} dx_idx_j}_{\circled{B}} + \\
    &\quad \underbrace{\iint \frac{\tau(x_i)(1-\omega(x_j))\mathbf{P}(x_i)\mathbf{P}(x_j)}{\mu_1(1-\mu_1)} \mathds{1}_{\{f(x_i)>f(x_j)\}} dx_idx_j - \iint \frac{\omega(x_i)\tau(x_j)\mathbf{P}(x_i)\mathbf{P}(x_j)}{\mu_1(1-\mu_1)} \mathds{1}_{\{f(x_i)>f(x_j)\}} dx_idx_j}_{\circled{C}}
\end{align*}

With some subtle rewrite, we can  express $\circled{A}$ as:
\begin{align*}
    \circled{A} &= \frac{\mu_0(1-\mu_0)}{\mu_1(1-\mu_1)}\iint \frac{\omega(x_i)}{\mu_0}\mathbf{P}(x_i)\frac{1-\omega(x_j)}{1-\mu_0}\mathbf{P}(x_j) \mathds{1}_{\{f(x_i)>f(x_j)\}} dx_idx_j \\
    &= \frac{\mu_0(1-\mu_0)}{\mu_1(1-\mu_1)}\iint \mathbf{P}(x_i|Y=1, T=0)\mathbf{P}(x_j|Y=0, T=0) \mathds{1}_{\{f(x_i)>f(x_j)\}}dx_idx_j \\
    &= \frac{\mu_0(1-\mu_0)\mathbb{E}\left[\text{AUC}_{\text{std}}(f)\right]}{\mu_1(1-\mu_1)}
\end{align*}

Leveraging the fact that the AUROC between two identical distributions is $\frac{1}{2}$, we can rewrite $\circled{B}$ as:

\begin{equation}
\begin{aligned}
     \circled{B} &= \frac{1}{\mu_1(1-\mu_1)}\iint \tau(x_i)\tau(x_j)\mathbf{P}(x_i)\mathbf{P}(x_j) \mathds{1}_{\{f(x_i)>f(x_j)\}} dx_idx_j \\
     &= \frac{1}{\mu_1(1-\mu_1)}\iint \tau(x_i)\tau(x_j)\mathbf{P}(x_i)\mathbf{P}(x_j) \frac{1}{2} dx_idx_j \\
     &=\frac{1}{2\mu_1(1-\mu_1)}\mathbb{E}[\tau(X)]\mathbb{E}[\tau(X)] \quad \text{($x_i$ and $x_j$ are independent)}\\
     &=\frac{\tau^2}{2\mu_1(1-\mu_1)}
\end{aligned}
\label{eq:b_derivation}
\end{equation}

Applying some algebraic manipulation, we can rewrite $\circled{C}$ as:
\begin{align*}
    \circled{C} = &\iint \frac{\tau(x_i)\mathbf{P}(x_i)\mathbf{P}(x_j)}{\mu_1(1-\mu_1)} \mathds{1}_{\{f(x_i)>f(x_j)\}} dx_idx_j - \iint \frac{\tau(x_i)\omega(x_j)\mathbf{P}(x_i)\mathbf{P}(x_j)}{\mu_1(1-\mu_1)} \mathds{1}_{\{f(x_i)>f(x_j)\}} dx_idx_j - \\
    &\iint \frac{\tau(x_i)\omega(x_j)\mathbf{P}(x_i)\mathbf{P}(x_j)}{\mu_1(1-\mu_1)} \left(1-\mathds{1}_{\{f(x_i)>f(x_j)\}}\right) dx_idx_j \quad \text{(Switch $x_i$ and $x_j$)} \\
    = &\underbrace{\iint \frac{\tau(x_i)\mathbf{P}(x_i)\mathbf{P}(x_j)}{\mu_1(1-\mu_1)} \mathds{1}_{\{f(x_i)>f(x_j)\}} dx_idx_j}_{\circled{C1}} - \underbrace{\iint \frac{\tau(x_i)\omega(x_j)\mathbf{P}(x_i)\mathbf{P}(x_j)}{\mu_1(1-\mu_1)}dx_idx_j}_{\circled{C2}}
\end{align*}

Applying some additional manipulations, we can rewrite $\circled{C1}$ as:
\begin{equation}
\begin{aligned}
    \circled{C1} = &\frac{1}{\mu_1(1-\mu_1)}\iint \mathbf{P}(x_i)\mathbf{P}(x_j) \tau(x_i)\mathds{1}_{\{f(x_i)>f(x_j)\}} dx_idx_j \\
    = &\frac{1}{\mu_1(1-\mu_1)}\int \mathbf{P}(x_i)\tau(x_i)\int\mathbf{P}(x_j) \mathds{1}_{\{f(x_i)>f(x_j)\}} dx_jdx_i \\
    = &\frac{1}{\mu_1(1-\mu_1)}\int \mathbf{P}(x_i)\tau(x_i)F_{f}(f(x_i)) dx_i \quad \text{($F_{f}(\cdot)$ denotes the CDF of model $f$'s output)} \\
    = &\frac{\mathbb{E}\left[\tau(X)F_f(f(X))\right]}{(\mu_1)(1-\mu_1)}
    = \frac{\mathbb{E}\left[\tau(X)\right]\mathbb{E}\left[F_{f}(f(X))\right] + \text{Cov}\left(\tau(X), F_{f}(f(X))\right)}{(\mu_1)(1-\mu_1)}  \\
    &= \frac{\tau}{2\mu_1(1-\mu_1)} + \frac{\text{Cov}\left(\tau(X), F_{f}(f(X))\right)}{\mu_1(1-\mu_1)}  \quad \text{(Expected CDF over different threshold is $\frac{1}{2}$)}
\end{aligned}
\label{eq:c1_derivation}
\end{equation}

Applying the fact that $\circled{C2}$ doesn't have terms depending on both $x_i$ and $x_j$, we obtain:
\begin{align*}
    \circled{C2} &= \frac{1}{\mu_1(1-\mu_1)}\iint \tau(x_i)\omega(x_j)\mathbf{P}(x_i)\mathbf{P}(x_j)dx_idx_j  \\
    &= \frac{\mathbb{E}\left[\tau(X)\right]\mathbb{E}\left[\omega(X)\right]}{\mu_1(1-\mu_1)} \quad \text{($x_i$ and $x_j$ are independent)} \\
    &= \frac{\tau\mu_0}{\mu_1(1-\mu_1)}
\end{align*}

Putting everything together, we have:
\begin{align*}
    \mathbb{E}\left[\text{AUC}_\text{igr}\left(f\right)\right] = &\frac{\mu_0(1-\mu_0)\mathbb{E}\left[\text{AUC}_{\text{std}}(f)\right]}{\mu_1(1-\mu_1)} -  \frac{\tau^2}{2\mu_1(1-\mu_1)} \\
    &+ \frac{\tau}{2\mu_1(1-\mu_1)} + \frac{\text{Cov}\left(\tau(X), F_{f}(f(X))\right)}{\mu_1(1-\mu_1)} - \frac{\tau\mu_0}{\mu_1(1-\mu_1)} \\
    = &\frac{\mu_0(1-\mu_0)\mathbb{E}\left[\text{AUC}_{\text{std}}(f)\right]}{\mu_1(1-\mu_1)} + \frac{\tau+2\text{Cov}\left(\tau(X), F_{f}(f(X))\right)-\tau^2-2\tau\mu_0}{2\mu_1(1-\mu_1)}
\end{align*}

Plug $\mathbb{E}\left[\text{AUC}_\text{igr}\left(f\right)\right]$ back in to the bias term, we have:
\begin{align*}
    \pi(\mathbb{E}&\left[\text{AUC}_{\text{std}}\left(f\right)\right] - \mathbb{E}\left[\text{AUC}_{\text{igr}}\left(f\right)\right]) 
    \\
    &= \pi\left(\frac{(\mu_1-\mu_1^2 - \mu_0+\mu_0^2)\mathbb{E}\left[\text{AUC}_{\text{std}}(f)\right]}{\mu_1(1-\mu_1)} - \frac{\tau+2\text{Cov}\left(\tau(X), F_{f}(f(X))\right)-\tau^2-2\tau\mu_0}{2\mu_1(1-\mu_1)}\right) \\
    &= \pi\left(\frac{\tau(1-\mu_0-\mu_1)\mathbb{E}\left[\text{AUC}_{\text{std}}(f)\right]}{\mu_1(1-\mu_1)} - \frac{\tau(1-\mu_0-\mu_1)0.5}{\mu_1(1-\mu_1)} - \frac{\text{Cov}\left(\tau(X), F_{f}(f(X))\right)}{\mu_1(1-\mu_1)}\right) \\
    &= \pi\frac{\tau(1-\mu_0-\mu_1)\left(\mathbb{E}\left[\text{AUC}_{\text{std}}(f)\right] - 0.5\right) - \text{Cov}\left(\tau(X), F_{f}(f(X))\right)}{\mu_1(1-\mu_1)}
\end{align*}

Lastly, we simplify the bias term by defining:
\begin{equation}
    \begin{split}
    \delta(f) &= \mathbb{E}\left[\text{AUC}_{\text{std}}(f)\right] - 0.5 \\
    \alpha &= \frac{\pi\tau(1-\mu_0-\mu_1)}{\mu_1(1-\mu_1)}
    \end{split}\quad
    \begin{split}
    \sigma(f) &= \text{Cov}\left(\tau(X), F_{f}(f(X))\right) \\
    \beta &= \frac{\pi}{\mu_1(1-\mu_1)}
    \end{split}
\label{eq:delta_sigma}
\end{equation}

Intuitively, $\delta(f)$ represents model $f$'s true AUROC improvement over a model that is predicting at random, and $\sigma(f)$ denotes the covariance between true individual-level intervention effect and model $f$'s CDF. 
$\alpha$ and $\beta$ are bounded rational numbers defined based on $\pi$, $\tau$, $\mu_0$, and $\mu_1$.
Specifically, $\alpha < 1$ and $\beta > 0$.
$\alpha$ is always less than $1$ because its numerator is composed of multiplication between $\pi\tau$ and $(1-\mu_0-\mu_1)$. These two terms are smaller than the two terms composed of the denominators: $\mu_1$ and $(1-\mu_1)$ respectively.
Because $\mu_1=\tau + \mu_0$ and $\pi \in [0, 1]$, we have: $\pi\tau < \mu_1$. Furthermore, because $\mu_0 \in (0, 1)$, we have $(1-\mu_0-\mu_1) < \mu_1(1-\mu_1)$.
Additionally, because both the numerator and denominator of $\beta$ are non-negative, we have $\beta>0$.

Lastly, substituting in the above notation, we have the final bias expression as:
\begin{align*}
    \text{Bias}\left(\text{AUC}_{\text{na\"ive}}(f)\right) = \alpha\delta(f) - \beta\sigma(f)
\end{align*}

\section{Bias of AUROC Estimated with All RCT Data}
\label{sec:appendix:all-expr}
We formally express the AUROC estimated using all the data from an RCT ($\mathbb{D}$) to estimate a model's AUROC as:
\begin{align*}
    \text{AUC}_{\text{all}}(f) = \text{AUC}\left(\mathbb{X}_0^+\cup \mathbb{X}_1^+, \mathbb{X}_0^-\cup \mathbb{X}_1^-, f\right)
\end{align*}

Applying the definition of AUROC and the assumed DGP, we can obtain the following expected $\text{AUC}_{\text{all}}(f)$:
\begin{align*}
    \mathbb{E}\left[\text{AUC}_{\text{all}}(f)\right] = \mathbb{E}\left[\mathds{1}_{\left\{f(x^+) > f(x^-)\right\}}\right], x^+ \sim \mathbf{P}(X | Y=1), x^- \sim \mathbf{P}(X | Y=0)
\end{align*}
where $\mathbf{P}(X | Y=1)$ and $\mathbf{P}(X | Y=0)$ can be further decomposed given randomized intervention:
\begin{align*}
    \mathbf{P}(X | Y=1) &= \mathbf{P}(X | Y=1, T=0)(1-\pi) + \mathbf{P}(X | Y=1, T=1)\pi \\
    \mathbf{P}(X | Y=0) &= \mathbf{P}(X | Y=0, T=0)(1-\pi) + \mathbf{P}(X | Y=0, T=1)\pi
\end{align*}

Then, substitute in the Bayes rule rewrite of the conditional probabilities from Equation~\ref{eq:proba_bayes_simple}, we have:
\begin{align*}
    \mathbf{P}(X | Y=1) &= \frac{\omega(X)}{\mu_0}\mathbf{P}(X)(1-\pi) + \frac{\omega(X)+\tau(X)}{\mu_1}\mathbf{P}(X)\pi \\
    \mathbf{P}(X | Y=0) &= \frac{1-\omega(X)}{1-\mu_0}\mathbf{P}(X)(1-\pi) + \frac{1-\omega(X)-\tau(X)}{1-\mu_1}\mathbf{P}(X)\pi
\end{align*}

Rewriting the expectation into integration form and substituting in the above expressions, we have:
\begin{align*}
    \mathbb{E}\left[\text{AUC}_{\text{all}}(f)\right] = \iint &\left(\frac{\omega(x_i)}{\mu_0}\mathbf{P}(x_i)(1-\pi) + \frac{\omega(x_i)+\tau(x_i)}{\mu_1}\mathbf{P}(x_i)\pi\right) \\
    &\left(\frac{1-\omega(x_j)}{1-\mu_0}\mathbf{P}(x_j)(1-\pi) + \frac{1-\omega(x_j)-\tau(x_j)}{1-\mu_1}\mathbf{P}(x_j)\pi\right)\mathds{1}_{\left\{f(x_i)>f(x_j)\right\}}dx_i dx_j
\end{align*}

Applying the linearity of integration, we have:
\begin{align*}
\mathbb{E}\left[\text{AUC}_{\text{all}}(f)\right] = 
&\iint \frac{\omega(x_i)(1 - \omega(x_j))}{\mu_0(1 - \mu_0)} \mathbf{P}(x_i)\mathbf{P}(x_j)(1 - \pi)^2 \, \mathds{1}_{\{f(x_i) > f(x_j)\}} \, dx_i \, dx_j \\
+ &\iint \frac{\omega(x_i)(1 - \omega(x_j) - \tau(x_j))}{\mu_0(1 - \mu_1)} \mathbf{P}(x_i)\mathbf{P}(x_j)(1 - \pi)\pi \, \mathds{1}_{\{f(x_i) > f(x_j)\}} \, dx_i \, dx_j \\
+ &\iint \frac{(\omega(x_i) + \tau(x_i))(1 - \omega(x_j))}{\mu_1(1 - \mu_0)} \mathbf{P}(x_i)\mathbf{P}(x_j)\pi(1 - \pi) \, \mathds{1}_{\{f(x_i) > f(x_j)\}} \, dx_i \, dx_j \\
+ &\iint \frac{(\omega(x_i) + \tau(x_i))(1 - \omega(x_j) - \tau(x_j))}{\mu_1(1 - \mu_1)} \mathbf{P}(x_i)\mathbf{P}(x_j)\pi^2 \, \mathds{1}_{\{f(x_i) > f(x_j)\}} \, dx_i \, dx_j
\end{align*}

Applying equation~\ref{eq:proba_bayes_simple}, we then rewrite the above as weighted averages of different AUROCs:
\begin{align*}
    \mathbb{E}\left[\text{AUC}_{\text{all}}(f)\right] = &(1-\pi)^2\mathbb{E}\left[\text{AUC}(\mathbb{X}_0^+, \mathbb{X}_0^-, f)\right] + (\pi-\pi^2)\mathbb{E}\left[\text{AUC}(\mathbb{X}_0^+, \mathbb{X}_1^-, f)\right] +\\
    &(\pi-\pi^2)\mathbb{E}\left[\text{AUC}(\mathbb{X}_1^+, \mathbb{X}_0^-, f)\right] + \pi^2\mathbb{E}\left[\text{AUC}(\mathbb{X}_1^+, \mathbb{X}_1^-, f)\right] \\
    = &\mathbb{E}\left[(1-\pi)^2\text{AUC}(\mathbb{X}_0^+, \mathbb{X}_0^-, f) + \pi^2 \text{AUC}(\mathbb{X}_1^+, \mathbb{X}_1^-, f) + \right.\\
    &\left.\quad (\pi-\pi^2)(\text{AUC}(\mathbb{X}_0^+, \mathbb{X}_1^-, f) + \text{AUC}(\mathbb{X}_1^+, \mathbb{X}_0^-, f))
    \right]
\end{align*}

\section{Derivation for NPW Augmented AUROC}
\label{sec:appendix:npw}

\subsection{Recovering Control Distributions with $\omega(X)$}
To recover the positive and negative individuals' sampling distribution under control, we can apply the Bayes' rule and the assumed DGP as Equation~\ref{eq:proba_bayes_simple} to obtain the following:
\begin{align*}
    \mathbf{P}(X | Y=1, T=0) = \frac{\omega(X)}{\mu_0}\mathbf{P}(X),\quad \mathbf{P}(X | Y=0, T=0) = \frac{1-\omega(X)}{1-\mu_0}\mathbf{P}(X)
\end{align*}

Because in RCT, intervention is randomly assigned, individuals in $\mathbb{D}_1$ represent samples from $\mathbf{P}(X)$. 
Therefore, we can simulate samples from the control distributions by weighting $\mathbb{D}_1$ following the above equation.
In practice, weighted sampling is inefficient; therefore, we instead compute the AUROC with $\hat{\omega}(X)$ as weighted AUROC, which is equivalent to sampling:
\begin{align*}
    \text{AUC}(\mathbb{X}_{\hat{\omega}(x)}^+, \mathbb{X}_{\hat{\omega}(x)}^-, f) = \frac{1}{\mu_0(1-\mu_0)}\sum_{\{x_i\mid t_i=1\}_{i=1}^n}\sum_{\{x_j\mid t_i=1\}_{j=1}^n}\omega(x_i)(1-\omega(x_j))\mathds{1}_{f(x_i)>f(x_j)}
\end{align*}

\subsection{Recovering Control Distributions with $\tau(X)$}

Also leveraging Equation~\ref{eq:proba_bayes_simple}, we can rewrite the control distributions in terms of their treatment distribution counterparts:
\begin{align*}
    \mathbf{P}(X | Y=0, T=0) &=\frac{1-\mu_1}{1-\mu_0}\mathbf{P}(X | Y=0, T=1) + \frac{\tau(X)}{1-\mu_0}\mathbf{P}(X) \\
    \mathbf{P}(X | Y=1, T=0) &=\frac{\mu_1}{\mu_0}\mathbf{P}(X | Y=1, T=1) - \frac{\tau(X)}{\mu_0}\mathbf{P}(X) 
\end{align*}

With the above formulation, the control distributions can also be recovered by aggregating the treatment distributions, $\mathbf{P}(X | Y=0, T=1)$ and $\mathbf{P}(X | Y=0, T=1)$, with the unconditional distribution of the target population, $\mathbf{P}(X)$.
In practice, we do not have access to either the treatment or unconditional distributions, but have samples from these distributions.
Aggregating samples from two distributions with weights that may be negative can be difficult to implement.
Therefore, we also leverage the weighted AUROC computation to simplify the implementation.

Then, we can express the expected AUROC by weighting samples following the equations above as:
\begin{align*}
    \mathbb{E}&\left[\text{AUC}\left(\mathbb{X}_{\hat{\tau}(X)}^+, \mathbb{X}_{\hat{\tau}(X)}^-, f\right)\right] = \iint \left(\frac{\mu_1}{\mu_0}\mathbf{P}(x_i | Y=1, T=1) - \frac{\tau(x_i)}{\mu_0}\mathbf{P}(x_i) \right) \\
    &\qquad\qquad\qquad\qquad\qquad\qquad\qquad\left(\frac{1-\mu_1}{1-\mu_0}\mathbf{P}(x_j | Y=0, T=1) + \frac{\tau(x_j)}{1-\mu_0}\mathbf{P}(x_j)\right)\mathds{1}_{\{f(x_i)>f(x_j)\}}dx_i dx_j \\
    = &\frac{\mu_1(1-\mu_1)}{\mu_0(1-\mu_0)}\iint \mathbf{P}(x_i \mid Y=1, T=1)\mathbf{P}(x_j \mid Y=0, T=1)\mathds{1}_{\{f(x_i)>f(x_j)\}}dx_idx_j + \\
    &\frac{\mu_1}{\mu_0(1-\mu_0)}\iint \mathbf{P}(x_i \mid Y=1, T=1)\tau(x_j)\mathbf{P}(x_j)\mathds{1}_{\{f(x_i)>f(x_j)\}}dx_idx_j - \\
    &\frac{1-\mu_1}{\mu_0(1-\mu_0)}\iint\tau(x_i)\mathbf{P}(x_i)\mathbf{P}(x_j \mid Y=0, T=1)\mathds{1}_{\{f(x_i)>f(x_j)\}}dx_idx_j - \\
    &\frac{1}{\mu_0(1-\mu_0)}\iint \tau(x_i)\mathbf{P}(x_i)\tau(x_j)\mathbf{P}\tau(x_j)\mathds{1}_{\{f(x_i)>f(x_j)\}}dx_idx_j \\
    = &\frac{\mu_1(1-\mu_1)}{\mu_0(1-\mu_0)}\mathbb{E}\left[\text{AUC}_{\mathbb{D}_1}(f)\right] - \frac{0.5\tau^2}{\mu_0(1-\mu_0)} + \\
    &\underbrace{\iint \left(\frac{\omega(x_i)+\tau(x_i)}{\mu_0(1-\mu_0)}\mathbf{P}(x_i)\tau(x_j)\mathbf{P}(x_j) - \tau(x_i)\mathbf{P}(x_i)\frac{1-\omega(x_j)-\tau(x_j)}{\mu_0(1-\mu_0)}\mathbf{P}(x_j)\right)\mathds{1}_{\{f(x_i)>f(x_j)\}}dx_idx_j}_{\circled{H}}
\end{align*}

We can further simplify the $\circled{H}$ term:
\begin{align*}
    \circled{H} = &\iint \frac{\omega(x_i)+\tau(x_i)}{\mu_0(1-\mu_0)}\mathbf{P}(x_i)\tau(x_j)\mathbf{P}(x_j)\mathds{1}_{\{f(x_i)>f(x_j)\}}dx_idx_j - \\
    &\iint \frac{\tau(x_i)}{\mu_0(1-\mu_0)}\mathbf{P}(x_i)\mathbf{P}(x_j)\mathds{1}_{\{f(x_i)>f(x_j)\}}dx_idx_j + \\
    &\iint \tau(x_i)\mathbf{P}(x_i)\frac{(\omega(x_j)+\tau(x_j))}{\mu_0(1-\mu_0)}\mathbf{P}(x_j)\mathds{1}_{\{f(x_i)>f(x_j)\}}dx_idx_j \\
    = &\iint \frac{\omega(x_i)+\tau(x_i)}{\mu_0(1-\mu_0)}\mathbf{P}(x_i)\tau(x_j)\mathbf{P}(x_j)\mathds{1}_{\{f(x_i)>f(x_j)\}}dx_idx_j - \\
    &\iint \frac{\tau(x_i)}{\mu_0(1-\mu_0)}\mathbf{P}(x_i)\mathbf{P}(x_j)\mathds{1}_{\{f(x_i)>f(x_j)\}}dx_idx_j + \\
    &\iint \frac{(\omega(x_i)+\tau(x_i))}{\mu_0(1-\mu_0)}\mathbf{P}(x_i)\tau(x_j)\mathbf{P}(x_j)(1-\mathds{1}_{\{f(x_i)>f(x_j)\}})dx_idx_j \\
    =& \iint \frac{(\omega(x_i)+\tau(x_i))}{\mu_0(1-\mu_0)}\mathbf{P}(x_i)\tau(x_j)\mathbf{P}(x_j)dx_idx_j - \\
    &\iint \frac{\tau(x_i)}{\mu_0(1-\mu_0)}\mathbf{P}(x_i)\mathbf{P}(x_j)\mathds{1}_{\{f(x_i)>f(x_j)\}}dx_idx_j \\
    =& \frac{\mu_1\tau}{\mu_0(1-\mu_0)} - \frac{\mathbb{E}\left[\tau(x)F_f(f(x))\right]}{\mu_0(1-\mu_0)}
\end{align*}

Substitute back in $\circled{H}$, we have:
\begin{align*}
    \mathbb{E}\left[\text{AUC}\left(\mathbb{X}_{\hat{\tau}(X)}^+, \mathbb{X}_{\hat{\tau}(X)}^-, f\right)\right] = \frac{\mu_1(1-\mu_1)\mathbb{E}\left[\text{AUC}_{\mathbb{D}_1}(f)\right]}{\mu_0(1-\mu_0)} + \frac{(\mu_1-0.5\tau)\tau}{\mu_0(1-\mu_0)} - \frac{\mathbb{E}\left[\tau(x)F_f(f(x))\right]}{\mu_0(1-\mu_0)}
\end{align*}

Replacing the expectation terms with their unbiased sample estimates, we have obtained the reformulated AUROC estimate:
\begin{align*}
    \text{AUC}\left(\mathbb{X}_{\hat{\tau}(X)}^+, \mathbb{X}_{\hat{\tau}(X)}^-, f\right) =  \frac{\hat{\mu}_1(1-\hat{\mu}_1)\text{AUC}_{\mathbb{D}_1}(f)}{\hat{\mu}_0(1-\hat{\mu}_0)} + \frac{(\hat{\mu}_1-0.5\hat{\tau})\hat{\tau}}{\hat{\mu}_0(1-\hat{\mu}_0)} - \frac{1}{\hat{\mu}_0(1-\hat{\mu}_0)|\mathbb{X}_1|}\sum_{x_i \in \mathbb{X}_1}\hat{\tau}(x)\hat{F}_f(f(x))
\end{align*}
where the empirical CDF, $\hat{F}_f(f(x))$, is defined as:
\begin{align*}
    \hat{F}_f(f(x_i)) = \frac{1}{|\mathbb{X}_1|}\sum_{x_j \in \mathbb{X}_1}\mathds{1}_{\{f(x_i) > f(x_j)\}}
\end{align*}

\section{Implementation Details}
All gradient boosted decision tree models for evaluation and nuisance parameter estimation are trained with randomized hyperparameter search.
Where the search space is shown in Table~\ref{tab:hyperparameters}.
All our experiments were conducted on a server with 128 CPU cores and 8 NVIDIA RTX 2080 GPUs for training and evaluation.
To implement our experiments and visualize the results, we used Python 3.11.5 and the following Python libraries: scikit-learn (version 1.5.0), numpy (1.26.4), pandas (2.2.2), xgboost (2.0.3), seaborn (0.13.2), and matplotlib (3.10.1).
Each experiment is repeated 100 times to quantify uncertainties.

\begin{table}[ht]
\centering
\begin{tabular}{|l|l|}
\hline
\textbf{Hyperparameter} & \textbf{Range/Distribution} \\ \hline
Learning Rate           & Uniform distribution between 0.001 and 0.1 \\ \hline
Max Depth               & Integer values between 5 and 30 \\ \hline
Max Leaves              & Integer values between 5 and 30 \\ \hline
Min Child Weight        & Uniform distribution between 0.1 and 10 \\ \hline
Gamma                   & Uniform distribution between 0.0 and 0.5 \\ \hline
Colsample by Tree       & Uniform distribution between 0.0 and 1.0 \\ \hline
Colsample by Level      & Uniform distribution between 0.0 and 1.0 \\ \hline
Colsample by Node       & Uniform distribution between 0.0 and 1.0 \\ \hline
Regularization Lambda   & Uniform distribution between 0.0 and 5.0 \\ \hline
Max Delta Step          & Uniform distribution between 0.0 and 10.0 \\ \hline
Scale Pos Weight        & Uniform distribution between 1 and 10 \\ \hline
Number of Estimators    & A set of values: 10, 100, 200, 300, 400, 500 \\ \hline
\end{tabular}
\caption{Hyperparameter search space}
\label{tab:hyperparameters}
\end{table}

\section{Additional Details about the Readmission Dataset}\label{appendix:readmission}
The readmission dataset comprises $1518$ samples from a target cohort eligible for the post-discharge intervention at Michigan Medicine, where the data RCT was conducted.
Among the dataset, $876$ (57.8\%) were randomized to the treatment group and $642$ (42.3\%) were randomized to the control group. 
We define the outcome of interest as the 30-day unplanned readmission outcome.
The 30-day unplanned readmission outcome was defined as an unscheduled hospital admission within 30 calendar days after their index hospital discharge. We determined whether an admission is scheduled based on their admission type in the electronic healthcare record (EHR).
Within the control group, 124  (19.3\%) experienced readmission; within the treatment group, 152  (17.4\%) experienced readmission.
We use $1518$ EHR features to model the nuisance parameters. 
These features include patient demographics, hospital utilization, and historical comorbidities.

\section{Additional Experimental Results}\label{appendix:exp}
Here, we provide additional experimental results under the synthetic and AMR-UTI datasets.

\begin{figure*}[!t]
  \centering
  \includegraphics[width=1.0\linewidth]{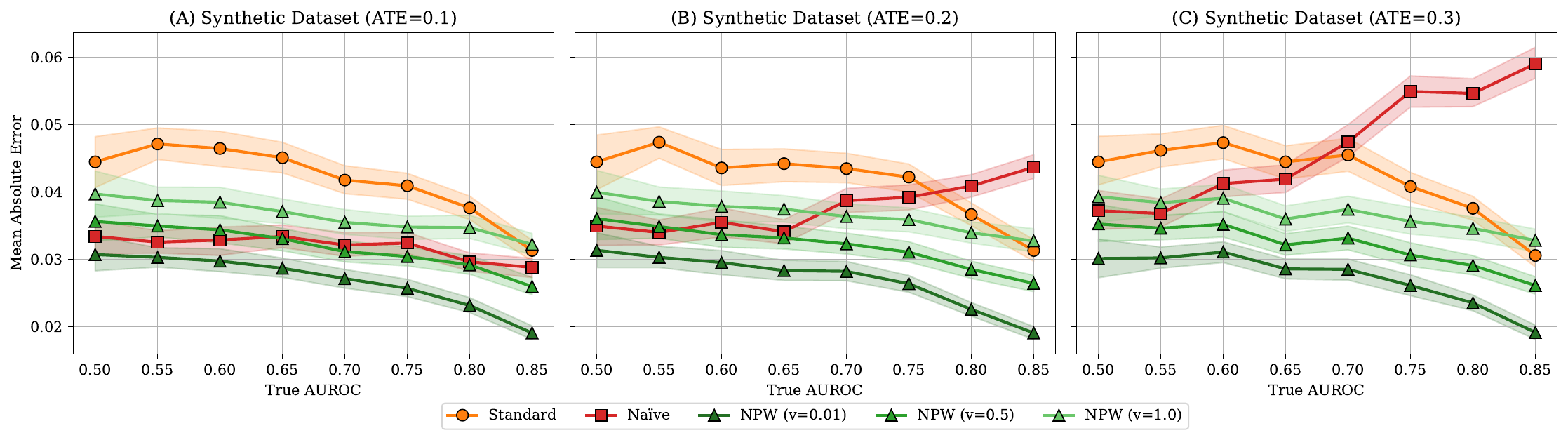}
  \captionof{figure}{Mean absolute errors (MAEs) of different AUROC estimations under various ATE and true model AUROCs with the synthetic dataset. 
  Our proposed NPW approach has consistent performance across different ATE settings.}
  \label{fig:sim_error_ATEs}
\end{figure*}
\subsection{Synthetic Experiments with Additional ATEs}\label{appendix:exp:sim_ates}
We compare NPW augmentation against the standard and \text{na\"ive} augmented AUROC under additional ATE settings.
As shown in Figure~\ref{fig:sim_error_ATEs}, NPW's performance is consistent across different ATE settings, indicating its unbiased nature.
It reduces MAE compared to standard AUROC estimates across all ATE settings.
Across all ATE settings, \text{na\"ive} augmented AUROC has increasing MAE as the model's true AUROC increases.
\text{na\"ive} augmented AUROC's performance degradation becomes more significant as the ATE increases. 
Both trends agree with our theoretical results, suggesting that the bias of \text{na\"ive} augmented is correlated with the model's true AUROC and interventions' ATE.

\subsection{Difference in \text{Na\"ive} Augmentation's Performance between Synthetic and AMR-UTI Dataset}\label{appendix:exp:naive_compare}

\begin{figure*}[!t]
  \centering
  \includegraphics[width=0.9\linewidth]{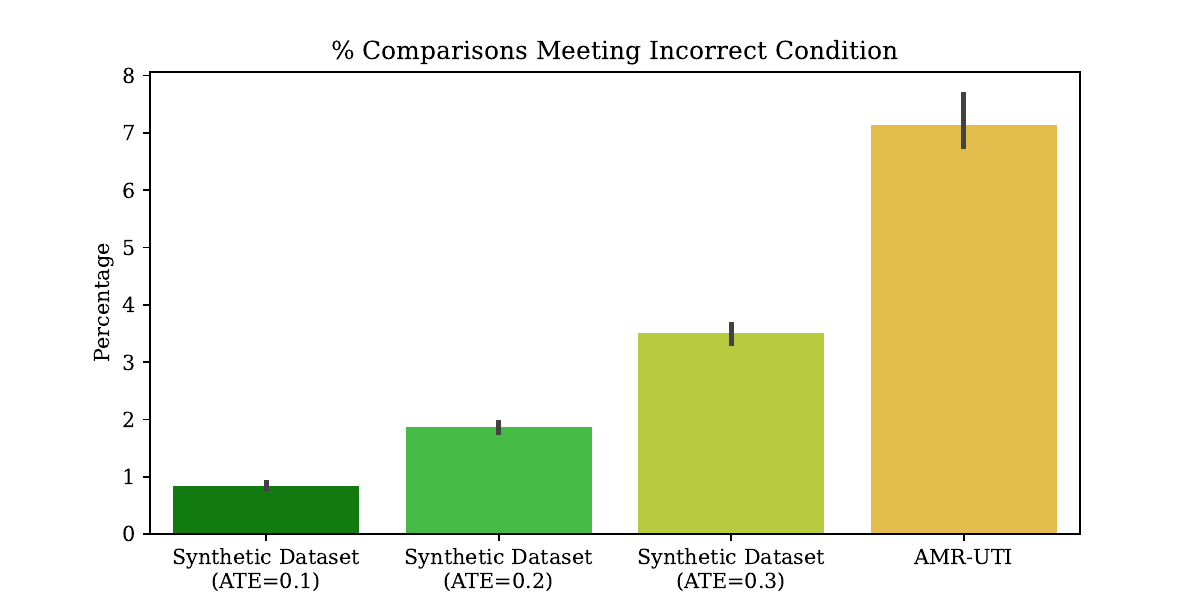}
  \captionof{figure}{Evaluating the percentage of all pairwise model comparisons in each dataset that meet the condition for \text{na\"ive} augmentation to derive incorrect model selection.
  The error bars represent the 95\% confidence interval.
  The average percentage of comparison meeting this condition consistently increases as the ATE increases in the synthetic dataset, agreeing with \text{na\"ive} augmentation's decreasing C-index performance.
  AMR-UTI has a significantly higher rate of incorrect comparison, explaining \text{na\"ive} augmentation's poor model selection performance in the AMR-UTI dataset.}
  \label{fig:data_naive_condition}
\end{figure*}

As shown in the experimental results in the main paper, \text{na\"ive} augmented AUROC improves model selection under the synthetic dataset but hurts model selection under the AMR-UTI dataset compared to the standard approach.
This difference is not due to a difference in the intervention's ATE, as the AMR-UTI dataset has an ATE of $0.0863$, similar to many settings in the synthetic dataset.
Leveraging Theorem 2, we explain this phenomenon by assessing all pair-wise model comparisons in each dataset on whether they meet the condition that \text{na\"ive} augmentation selects the incorrect model.
As shown in Figure~\ref{fig:data_naive_condition}, among synthetic datasets with different ATEs, the percentage of comparisons meeting the incorrect condition consistently increases as the ATE increases, explaining \text{nai\"ve} augmentation's decreasing performance as ATE increases.
Comparing the AMR-UTI dataset against the synthetic datasets, the same percentage significantly increases, resulting in \text{na\"ive} augmentation's poor C-index performance in the AMR-UTI dataset. This further highlights the importance of using NPW for unbiased augmentation.

\ifdefined\isMain
\else
  \end{document}
\fi

\end{document}